\documentclass{article}

\usepackage{arxiv}
\usepackage[T1]{fontenc}
\usepackage{bbm}
\usepackage{amsmath}
\usepackage{todonotes}
\usepackage{comment}
\usepackage{multirow}
\usepackage{graphicx}
\usepackage{hyperref}

\newcommand{\fontMetric}{\textsl}
\newcommand{\fontMethod}{\text}

\newcommand{\clarc}{\fontMethod{ClArC}}
\newcommand{\rrclarc}{\fontMethod{RR-ClArC}}
\newcommand{\aclarc}{\fontMethod{A-ClArC}}
\newcommand{\savaniaft}{\fontMethod{SavaniAFT}}
\newcommand{\zhangAL}{\fontMethod{ZhangAL}}
\newcommand{\thropt}{\fontMethod{ThrOpt}}
\newcommand{\Vanilla}{\fontMethod{Vanilla}}

\renewcommand{\shorttitle}{Investigating the Relationship Between Debiasing and Artifact Removal}

\hypersetup{
pdftitle={Investigating the Relationship Between Debiasing and Artifact Removal using Saliency Maps},
pdfsubject={},
pdfauthor={Lukasz Sztukiewicz, Ignacy Stepka, Michal Wilinski, Jerzy Stefanowski},
pdfkeywords={Deep Learning, Machine Learning, Fairness, Saliency Map, XAI},
}

\title{Investigating the Relationship Between Debiasing and Artifact Removal using Saliency Maps}

\begin{document}
\author{
  Lukasz Sztukiewicz$^{\ast}$ \quad Ignacy St\k{e}pka$^{\ast}$ \quad Micha\l{} Wili\'{n}ski$^{\ast}$ \quad Jerzy Stefanowski \\
  Institute of Computing Science, Poznan University of Technology
}

\maketitle

\begin{abstract}
The widespread adoption of machine learning systems has raised critical concerns about fairness and bias, making mitigating harmful biases essential for AI development. In this paper, we investigate the relationship between debiasing and removing artifacts in neural networks for computer vision tasks. First, we introduce a set of novel XAI-based metrics that analyze saliency maps to assess shifts in a model’s decision-making process. Then, we demonstrate that successful debiasing methods systematically redirect model focus away from protected attributes. Finally, we show that techniques originally developed for artifact removal can be effectively repurposed for improving fairness. 
These findings provide evidence for the existence of a bidirectional connection between ensuring fairness and removing artifacts corresponding to protected attributes.  

\end{abstract}

\section{Introduction}
\renewcommand{\thefootnote}{$\ast$}
\footnotetext{Authors contributed equally, and the order has been randomized.}
\renewcommand{\thefootnote}{\arabic{footnote}}

Machine learning (ML) systems are becoming widespread across numerous application domains. However, their adoption raises concerns about perpetuating harmful biases and creating discriminatory systems~\cite{Buyl_2024_inherent_limit}. 
This problem has been noticed not only by practitioners but also  policymakers, resulting in regulatory efforts \cite{wachter2018}, which underscore the importance of fairness in ML. 
Machine learning fairness refers to the principle of ensuring that algorithmic decisions do not produce biased or discriminatory outcomes across different groups.
Neural networks, especially in computer vision applications, present unique challenges for fairness assessment and bias mitigation \cite{tianImageFairnessDeep2022a}.
Unlike tabular data, where features are explicitly defined, images lack semantic meaning at the raw pixel level. To gain predictive power, models learn to extract high-level semantic features.
This learned featurization becomes problematic when dealing with protected attributes -- high-level features (concepts), such as gender, hair color or race, consisting of various pixel combinations.
Neural networks are known to develop internal representations that encode not only useful high-level features but also harmful biases \cite{bolukbasi_2016}.
For example, in the CelebA dataset \cite{celeba}, wearing a necktie is highly correlated with the male gender, and can be used as a proxy feature to infer gender, thus creating potential unintended pathways for discrimination.

To address discrimination and ensure fairness in ML models, various approaches have been proposed \cite{mehrabi_2021_survey,Caton2024}.
However, while existing debiasing methods generally improve fairness metrics, they often fail to explicitly address harmful biases encoded in models' internal representations.
To this end, we examine the relationship between successful fairness improvement and removal of harmful biases from these representations.
We propose new metrics that quantify the properties of saliency maps given a region of interest, and capture the extent to which biases are removed from the model's decision-making process.

Our findings provide evidence that effective debiasing methods redirect the model's focus away from protected attributes, while explicitly optimizing only the fairness criterion.
Furthermore, we observe that techniques originally developed for artifact removal, such as the family of \clarc\ methods \cite{dreyer2024hope}, also optimize fairness even though their explicit goal is to remove the designated artifact. 
These findings point to the existence of an inherent relationship between improving fairness and steering the saliency away from the protected attributes.

\section{Related Work}
\label{sec:related-work}

\textbf{Debiasing methods} are an active area of research, usually in the context of tabular data, with a vast landscape of methods applied at various stages of model development \cite{Caton2024}. The methods employed in our study represent approaches to debiasing in a post-hoc manner, that is, after a model is trained, within a binary classification setup. In our work, we consider three groups of methods. The first group consists of simple threshold optimizers, represented in our experiments by \thropt \cite{hardt2016equality}. The second group focuses on approaches that optimize fairness with adversarial fine-tuning and is represented by \zhangAL \cite{zhang_2018_adversarialbias} and \savaniaft \cite{Savani_2020_intra}. Finally, the third group focuses on concept-based interventions (artifact removal), exemplified by \clarc\ variants \cite{anders2022finding,dreyer2024hope}, which operate directly on the model's internal representations utilizing Concept Activation Vectors (CAVs) through interventions in activation space.

\textbf{Saliency maps} are explainable AI methods that provide insights into model decision-making process by highlighting regions of input data that influence predictions. 
These techniques can generally be categorized into gradient-based \cite{sundararajan2017axiomatic,molnar2022} and relevance-based methods \cite{bach2015pixel}.
Integrated Gradients (IG) \cite{sundararajan2017axiomatic} attributes predictions to input features by integrating gradients along a path from a baseline to the input, satisfying important axioms, including sensitivity and implementation invariance. 
Layer-wise Relevance Propagation (LRP) \cite{bach2015pixel} employs a different approach based on a conservation principle, where relevance scores are propagated backward through the network layers while maintaining a constant sum. To improve the faithfulness of our study, we conducted experiments with multiple saliency map methods, each providing a different perspective on model predictions and associated limitations \cite{rudin2019stop}.

\textbf{Quantitative evaluation} of saliency maps is crucial for assessing whether models make decisions based on appropriate features rather than biased artifacts or protected attributes. Early approaches, such as the inside-outside ratio \cite{kohlbrenner2020towards,Bach2015AnalyzingCF}, established a foundation by quantifying the relevance contained within a bounding box relative to the relevance outside it. This concept has been further developed as part of the Quantus toolbox \cite{hedstrom2023quantus}, which provides a framework for evaluating explanations through various localization metrics. Motzkus et al. \cite{motzkus2024locally} advanced this approach by adapting the inside-outside metric to compute the ratio of positively attributed relevance within a binary class mask to the overall positive relevance, specifically focusing on the context of individual concepts.

\section{Metrics for Saliency Maps}
\label{sec:metrics} 

In this section, we present metrics designed to quantify the importance of protected attributes in the model's decision-making process. Our focus is specifically on localized features that can be roughly bounded by rectangular regions of interest (ROIs). These metrics evaluate whether an ROI plays an important role in the model's reasoning by analyzing saliency maps. In principle, they can be used with any standard saliency map generation method that suits the practical needs of an application.

To establish our framework, we define several key components. Image $P$ is a 2D array with $p_{ij}$ representing the intensity (or relevance) of the pixel $(i,j)$. Within this image, we consider a 2D array (ROI) $R$ such that $|R| < |P|$.

\paragraph{Rectangle Relevance Fraction (RRF)} provides a direct measure of the ROI's importance in the context of the model's prediction by calculating what percentage of the total relevance falls within the region.

\begin{equation}
    \mathbf{RRF} = \frac{ \sum_{(i,j) \in R} p_{ij}}{ \sum_{(i,j) \in P }p_{ij}}
\end{equation}

 It aids in understanding the relative ROI's contribution to the overall decision-making process of the model.

\paragraph{Average Difference in Region (ADR)} provides a direct measure of how the saliency values within the ROI change after debiasing. It is defined as:

\begin{equation}
    \mathbf{ADR} = \frac{1}{\vert R \vert}\sum_{(i,j) \in R} p^\text{v}_{ij} - p^\text{d}_{ij} 
\end{equation}
where $p^\text{v}_{ij}$ and $p^\text{d}_{ij}$ represent pixel intensities in \Vanilla\ (corresponding to the base model) and debiased saliency maps, respectively. A positive ADR value indicates that \Vanilla\ generally assigned higher importance to pixels within the ROI compared to the debiased model, suggesting a successful reduction in the model's reliance on these features.

\paragraph{Decreased Intensity Fraction (DIF)} quantifies the proportion of pixels within the ROI that show reduced importance after debiasing. Specifically, it calculates the fraction of pixels where the debiased model shows lower saliency values compared to the \Vanilla\ model. It is defined as:
\begin{equation}
    \mathbf{DIF} = \frac{1}{\vert R \vert} \sum_{(i,j) \in R} \mathbbm{1}_{\{p^\text{d}_{ij} < p^\text{v}_{ij}\}}
\end{equation}
DIF provides insight into how widespread the changes are within the ROI, complementing the ADR's measurement of average change magnitude.

\paragraph{Rectangle Difference Distribution Testing (RDDT)} metric assesses whether \Vanilla\ assigns higher importance to pixels within the ROI compared to the debiased model. For each image, we compute the difference between the mean intensities of vanilla and debiased saliency maps within the ROI:
\begin{equation}
d = \mu_{\text{vanilla}} - \mu_{\text{debiased}}
\end{equation}
where $\mu_{\text{vanilla}}$ and $\mu_{\text{debiased}}$ represent the mean pixel intensities within the ROI for the \Vanilla\ and debiased models respectively. We then perform a one-sample t-test on these differences across with $H_0: \mu_d = 0$ and $H_1: \mu_d > 0$.
The test returns 1 if $p < 0.01$, indicating statistically significant evidence that the \Vanilla\ model assigns a higher importance to the ROI than the debiased model, and 0 otherwise.

\section{Experiments}

In the experiments below, we aim to explore the following two research questions. 
\textbf{RQ1:} Is there a bidirectional relationship between shifting the importance of pixels in the saliency map out of the ROI and optimizing fairness metrics? 
\textbf{RQ2:} Are debiasing methods capable of decreasing the saliency within ROI w.r.t. a standard end-to-end trained \Vanilla\ model?

For our experiments, we utilize methods detailed in Sec.~\ref{sec:related-work}, implemented within the DetoxAI library \cite{detoxai2025}. We compute metrics and generate visualizations using LRP and Integrated Gradients. To ensure reproducibility, we have open-sourced a GitHub repository containing the relevant implementations \footnote{\url{https://github.com/DetoxAI/saliency-fairness-metrics}}.

The experimental procedure begins by fine-tuning a pre-trained ResNet-50 \cite{Resnet} on the target task's training set, yielding our \Vanilla\ model. This fine-tuning uses a batch size of 128, the Adam optimizer, and a learning rate of $3 \cdot 10^{-4}$ for a single epoch. Subsequently, we apply the considered debiasing methods using a disjoint hold-out (debias) set. Finally, we evaluate the resulting models on a test set, calculating prediction performance, fairness, and our proposed metrics. Notably, both the training and debias datasets maintain the same \textit{protected attribute-target} (PA-T) correlation, reflecting a common practical scenario where the split strategy is fixed. In contrast, the test set intentionally balances the PA-T correlation to systematically assess predictive performance (Accuracy) and fairness (EqualizedOdds)~\cite{brzezinski2024properties}.

\subsection{Qualitative assessment}
\begin{figure}[tb]
    \centering
    \includegraphics[width=0.95\textwidth]{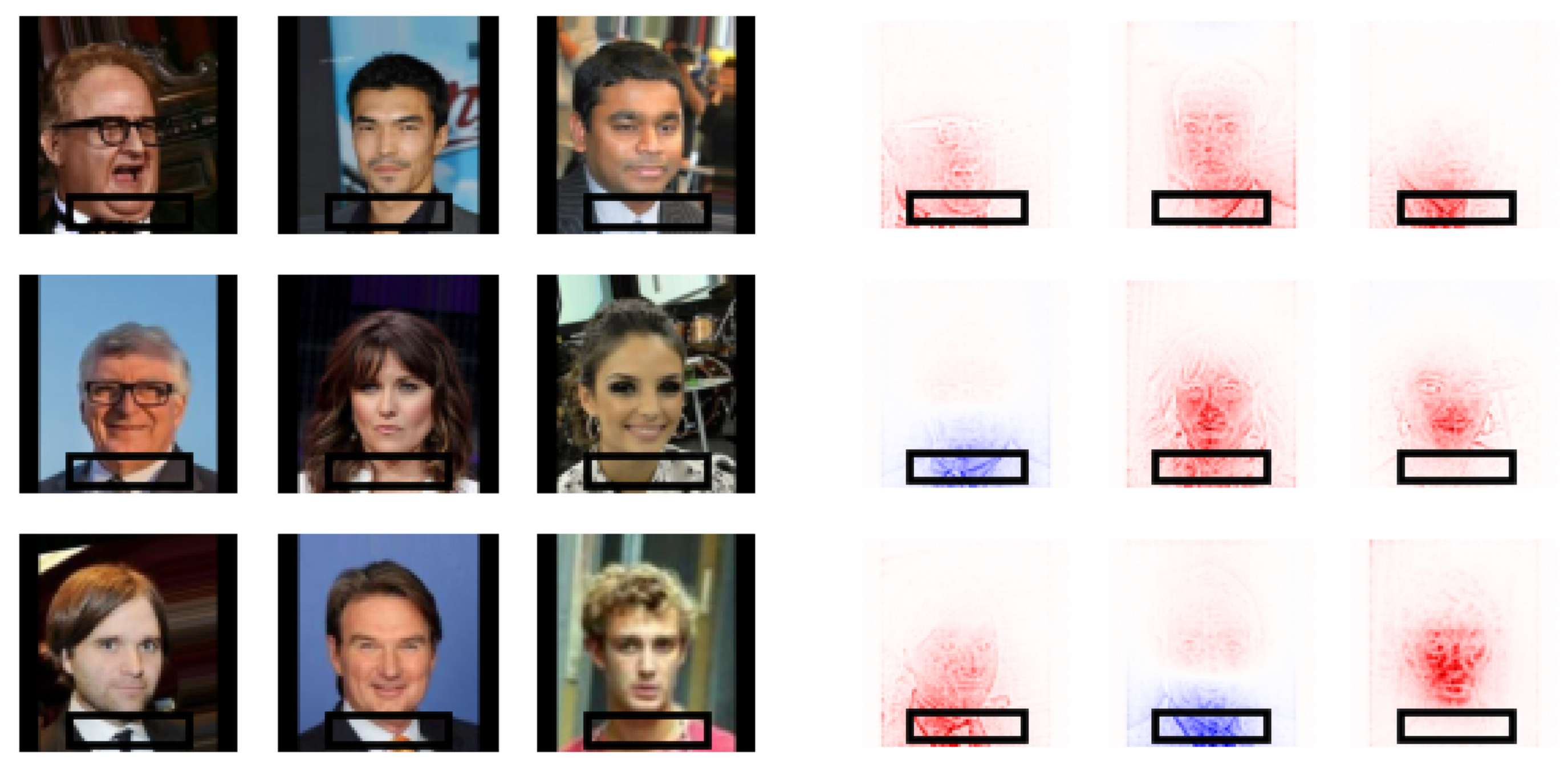}
    \hfill
    \includegraphics[width=0.038\textwidth]{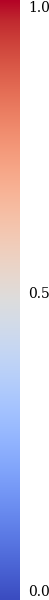}
    \caption{The left panel shows raw images, and the right panel displays corresponding LRP saliency maps. In the saliency maps, red hues indicate positive relevance towards the true class, while blue hues indicate negative contributions.}
    \label{fig:enter-label}
\end{figure}

\begin{figure}[tb]
    \centering
    \begin{minipage}[c][6.5cm]{0.08\textwidth}
        \vspace{0.45cm}
        \scriptsize{PA=0 T=0} \\
        \vfill
        \scriptsize{PA=1 T=0} \\
        \vfill
        \scriptsize{PA=0 T=1} \\
        \vfill
        \scriptsize{PA=1 T=1}
        \vfill
    \end{minipage}\hfill
    \begin{minipage}{0.124\textwidth}
        \centering
        \scriptsize{Image} \\
        \vspace{5pt}
        \includegraphics[width=\textwidth]{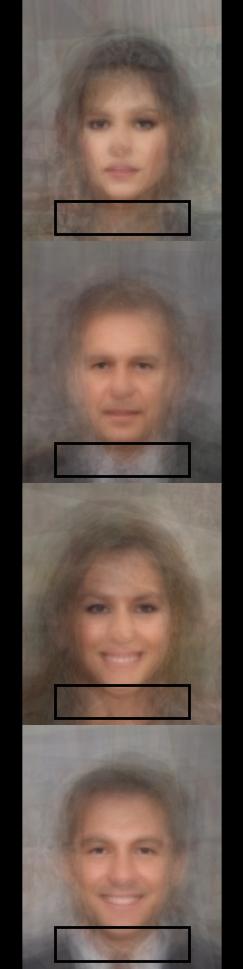}
    \end{minipage}\hfill
    \begin{minipage}{0.124\textwidth}
        \centering
        \scriptsize{\Vanilla}\\
        \vspace{5pt}
        \includegraphics[width=\textwidth]{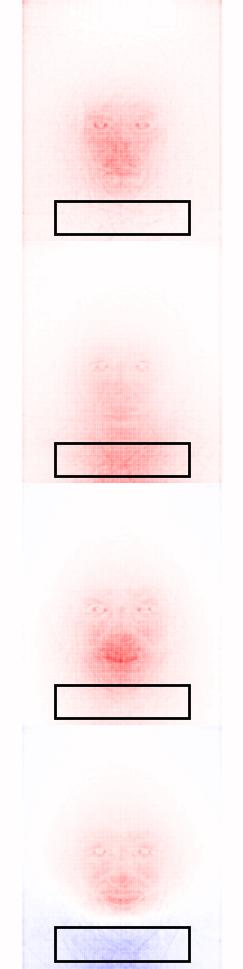}
    \end{minipage}\hfill
    \begin{minipage}{0.124\textwidth}
        \centering
        \scriptsize{\thropt}\\
        \vspace{5pt}
        \includegraphics[width=\textwidth]{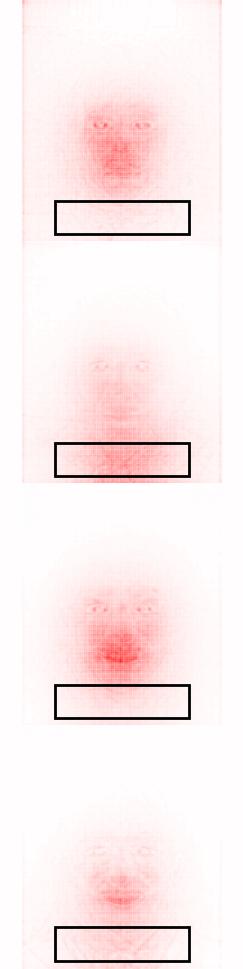}
    \end{minipage}\hfill
    \begin{minipage}{0.124\textwidth}
        \centering
        \scriptsize{\savaniaft}\\
        \vspace{5pt}
        \includegraphics[width=\textwidth]{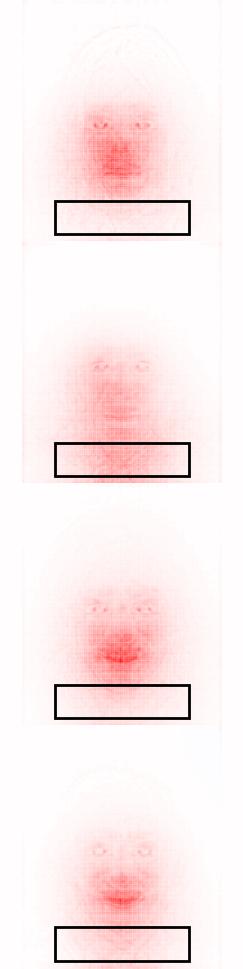}
    \end{minipage}\hfill
    \begin{minipage}{0.124\textwidth}
        \centering
        \scriptsize{\zhangAL}\\
        \vspace{5pt}
        \includegraphics[width=\textwidth]{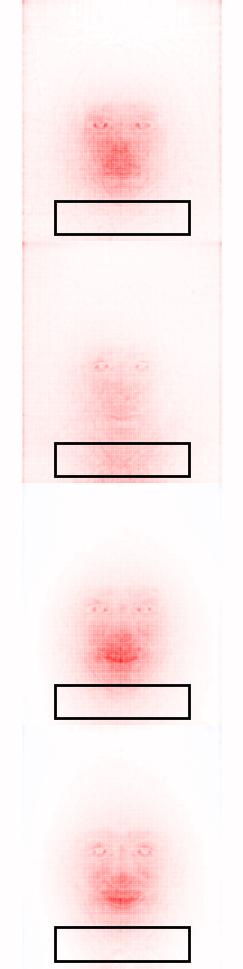}
    \end{minipage}\hfill
    \begin{minipage}{0.124\textwidth}
        \centering
        \scriptsize{\aclarc}\\
        \vspace{5pt}
        \includegraphics[width=\textwidth]{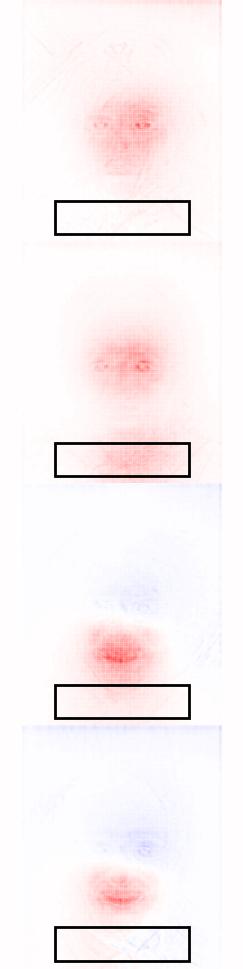}
    \end{minipage}\hfill
    \begin{minipage}{0.124\textwidth}
        \centering
        \scriptsize{\rrclarc}\\
        \vspace{5pt}
        \includegraphics[width=\textwidth]{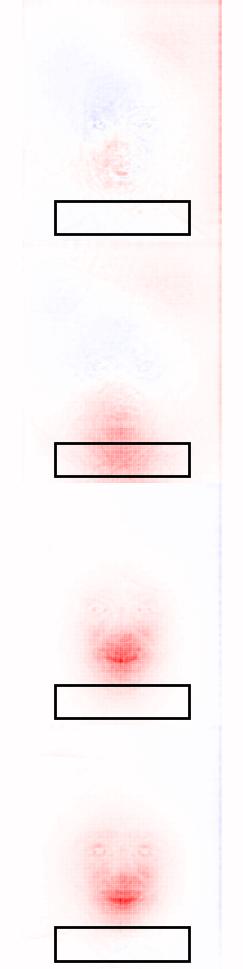}
    \end{minipage}
    \caption{ LRP saliency maps, averaged over a batch of 128 images and grouped by protected attribute (WearingNecktie) and target (Smiling) combinations. PA=1 indicates WearingNecktie, T=1 indicates Smiling.} 
    \label{fig:lrp_saliency}
\end{figure}

We perform a qualitative assessment of the debiasing by inspecting the relevancy maps before and after applying different debiasing methods. Fig.~\ref{fig:lrp_saliency} presents LRP saliency maps for images aggregated by PA-T combinations, where the protected attribute is \textit{WearingNecktie} and the target attribute is \textit{Smiling}. The black rectangles highlight the ROI roughly corresponding to the necktie area (see Fig.~\ref{fig:enter-label}).

Several key observations can be made from these visualizations.
The \Vanilla\ model (second column) shows considerable attention to the necktie region, particularly for the (PA=1, T=0) combination, indicating that the model has learned to associate the necktie area with its predictions. Interestingly, for the (PA=1, T=1) combination (bottom row), the necktie area shows strong negative relevance (blue), suggesting the model uses this feature to make negative predictions about smiling.

Simple threshold optimization (\thropt) does not substantially alter the saliency patterns compared to \Vanilla, maintaining similar attention to the necktie area. This suggests that merely adjusting classification thresholds does not change the underlying reasoning of the model.
Adversarial fine-tuning methods (\savaniaft\ and \zhangAL) show modest reductions in the attention to the ROI but largely preserve the overall saliency patterns of the \Vanilla\ model.
The \clarc-based methods show the most noticeable shifts. \aclarc\ reduces the saliency in the necktie region across all PA-T combinations, redirecting attention to facial features, relevant to the \textit{Smiling} attribute. 
\rrclarc\ shows the most visible improvements, excluding the second row, almost completely eliminating the relevance from ROI. 
These observations suggest that, while all debiasing methods may improve fairness metrics, they differ in how they alter the model's underlying decision-making process. Methods from the \clarc\ family most effectively redirect the model's attention away from the protected attribute region.

\subsection{Quantitative experiments}
While the CelebA dataset exhibits inherent attribute correlations, we artificially enforced specific PA-T correlations in our experimental framework to amplify the biases. This was done by rebalancing the dataset by undersampling attribute combinations to control their correlation with the target, as captured by Yule's correlation coefficient $\phi$. 

\begin{figure}[tb]
    \centering
    \includegraphics[width=\textwidth]{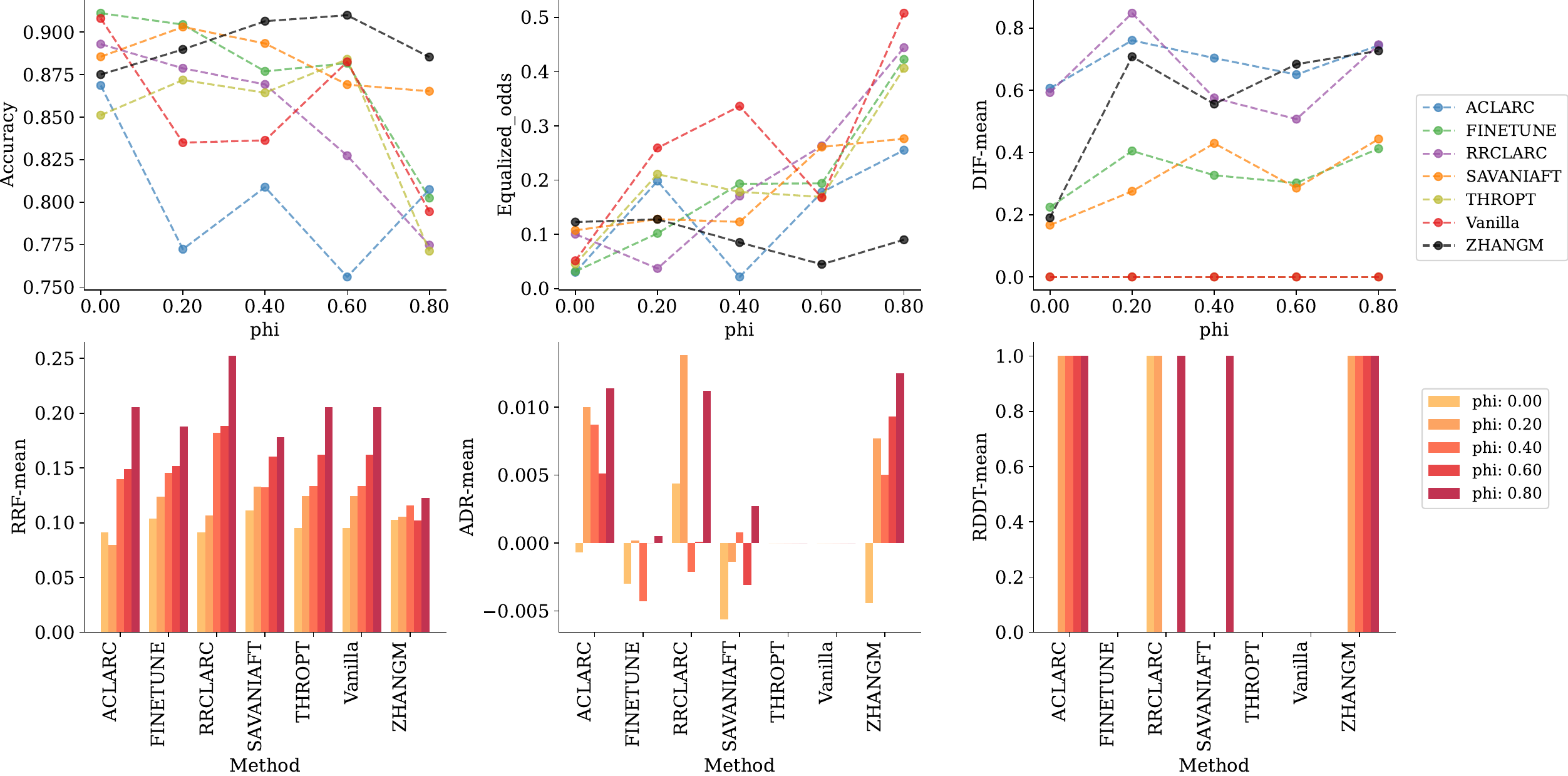}
    \caption{Quantitative metrics for WearingNecktie-Smiling PA-T classification task, measured on saliency maps generated with LRP. Metrics in the upper row are supposed to be minimized, while in the lower row, maximized.}
    \label{fig:lrp-mixed}
\end{figure}

In this experiment, we considered two PA-T combinations: \textit{WearingHat--Smiling} and \textit{WearingNecktie--Smiling}, using saliency maps generated with LRP \cite{bach2015pixel} and IntegratedGradients \cite{sundararajan2017axiomatic}. However, in the following, we only report the results for LRP and \textit{WearingNecktie--Smiling} combination (in Fig.~\ref{fig:lrp-mixed}), while we move the rest to the Appendix, because the conclusions from all experiment variants are the same. In these plots, we report metrics from Sec.~\ref{sec:metrics} along with EqualizedOdds calculated as: 
\begin{equation}
   \fontMetric{EqualizedOdds} = \max\big(\lvert \fontMetric{TPR}_{PA=1} - \fontMetric{TPR}_{PA=0} \rvert, \lvert \fontMetric{FPR}_{PA=1} - \fontMetric{FPR}_{PA=0} \rvert\big)
\end{equation}
where TPR and FPR stand for true and false positive rates respectively, and $PA=0$, $PA=1$ protected attribute value assignments.


First, it is clear that as $\phi$ increases, all methods achieve a higher EqualizedOdds value, which indicates more bias in their predictions. The best performing method for this metric is \zhangAL, which optimizes it directly internally. However, most methods decrease the EqualizedOdds score w.r.t. \Vanilla's, confirming that they are effective.

\thropt, a post-hoc classification threshold optimization method, does not shift the relevancy in or out of the ROI. Its bars are empty for ADR and RDDT and equal to \Vanilla\ on DIF and RRF, indicating that no change in the saliency maps was recorded. This is expected since no changes are made to the reasoning process with this method.  This method causes decrease in accuracy as the correlation grows larger.

\savaniaft\ and \zhangAL\ both perform well across most metrics. \zhangAL\ scores remarkably well in saliency map-based metrics. It lowers all but one metric value in the first row of the plot, showing that it moves the saliency out of ROI. As correlation grows, accuracy of the model also grows. In addition, it also scores visibly well on the metrics in the lower row, which measure the improvement over the \Vanilla\ model within the ROI. This provides evidence that optimizing with a fairness-oriented objective as a fine-tuning step can significantly shift the model's reasoning process.

In particular, \rrclarc\ and \aclarc\ methods do not optimize any fairness objective. Yet, they effectively debias the model (as captured by EqualizedOdds) and significantly shift model relevancy within the ROI. Both score high at DIF and ADR, and often appear on RDDT (the more bars the better). Regarding attention outside the ROI, they tend to lower the RRF with respect to \Vanilla, which suggests that more attention is given to features outside the ROI, - the desired outcome. Both methods cause decrease in accuracy.

\section{Conclusion}

Experiments show that effective debiasing methods decrease saliency within the ROI compared to the \Vanilla\ model, which positively answers RQ2. 
Both qualitative and quantitative analyses reveal that while threshold optimization (\thropt) produces no changes in saliency maps, fine-tuning-based approaches yield significant improvements. 
Notably, \zhangAL\ and \savaniaft\ and \clarc-based methods (\aclarc\ and \rrclarc) redirect the attention away from protected features towards task-relevant features such as facial expressions for smile detection. For the latter, the saliency redirection is stronger while achieving competitive EqualizedOdds, despite not directly optimizing any fairness objective.

These findings provide evidence for a bidirectional relationship between shifting pixel importance in saliency maps away from regions of interest and optimizing fairness metrics, validating the premise of RQ1. 
They confirm that methods that effectively redirect model attention away from protected attributes tend to score better on EqualizedOdds, and vice versa. 

We believe that this research provides useful evidence for further work on fairness methods, which could adapt concept removal methods directly in the field of fair machine learning. 

\bibliographystyle{splncs04}
\bibliography{mybibliography}

\clearpage

\section*{Appendix}
\label{appendix}

\vfill

\begin{figure}[htb]
    \centering
    \includegraphics[width=\textwidth]{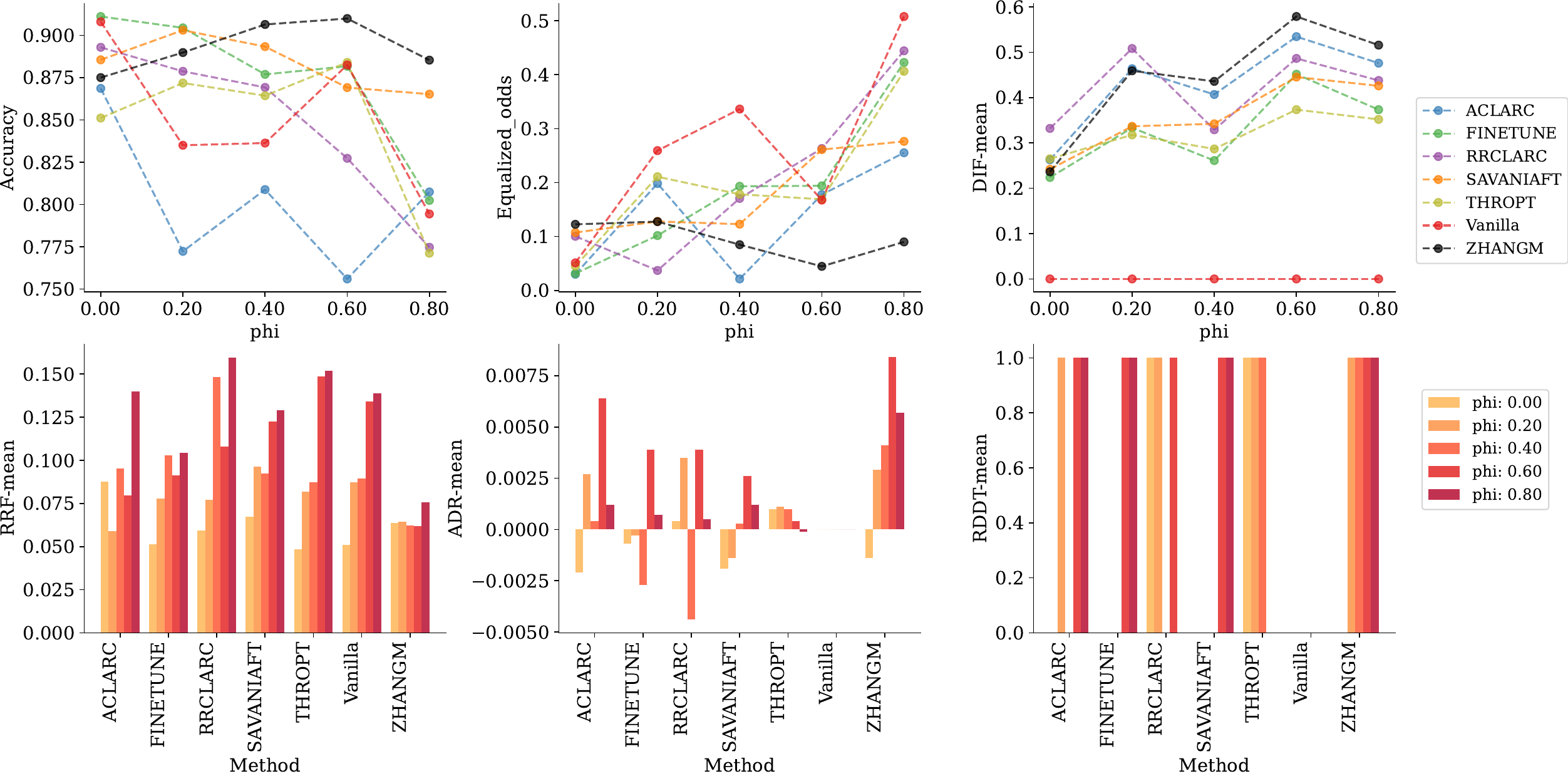}
    \caption{Metric values for the \texttt{IG} attributions and \textit{WearingNecktie} protected attribute.}
    \label{fig:IntegratedGradients-WearingNecktie}
\end{figure}

\begin{figure}[htb]
    \centering
    \includegraphics[width=\textwidth]{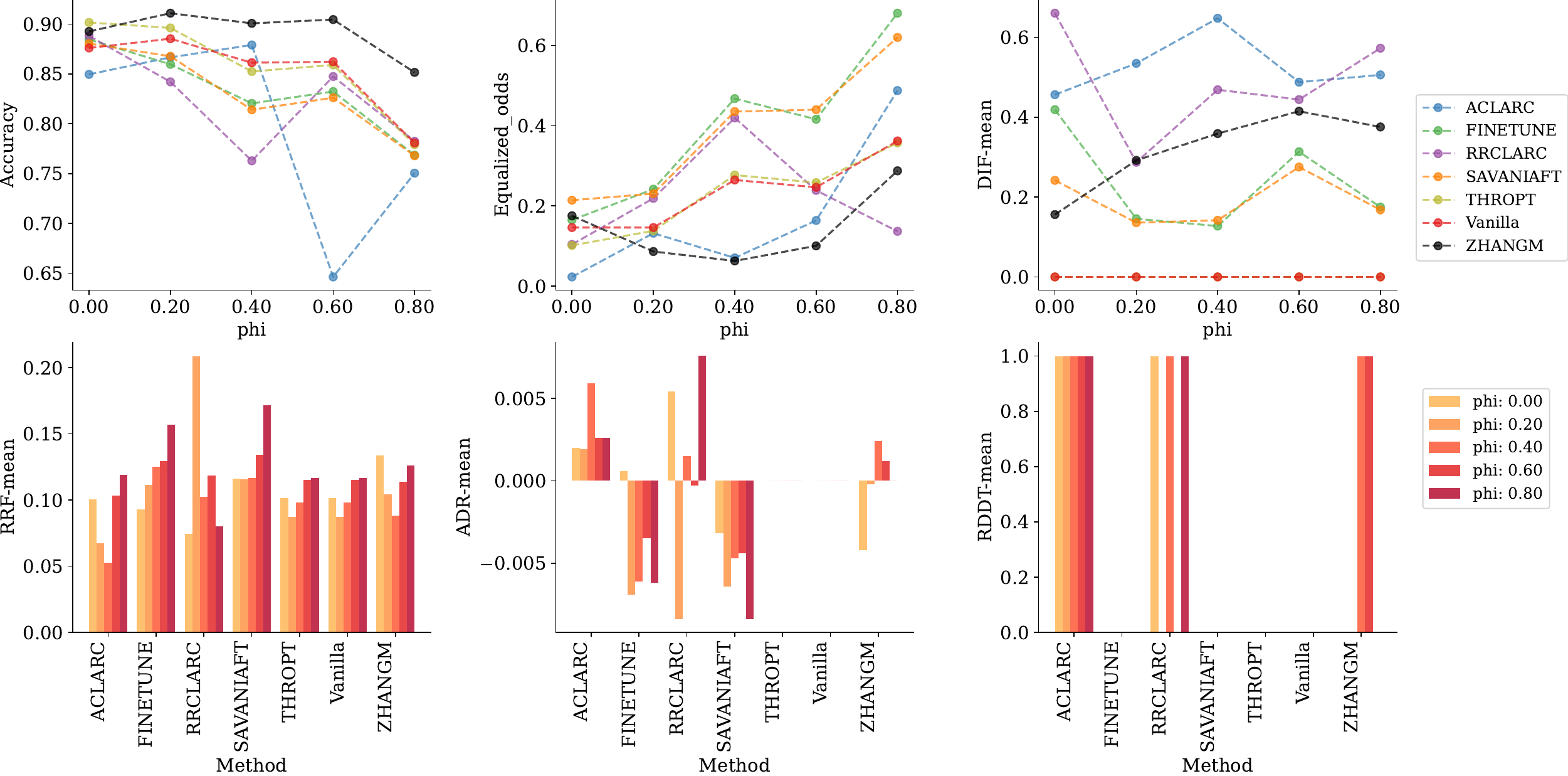}
    \caption{Metric values for the \texttt{LRP} attributions and \textit{WearingHat} protected attribute}
    \label{fig:LRP-mixedplot-WearingHat}
\end{figure}

\begin{figure}[htb]
    \centering
    \includegraphics[width=\textwidth]{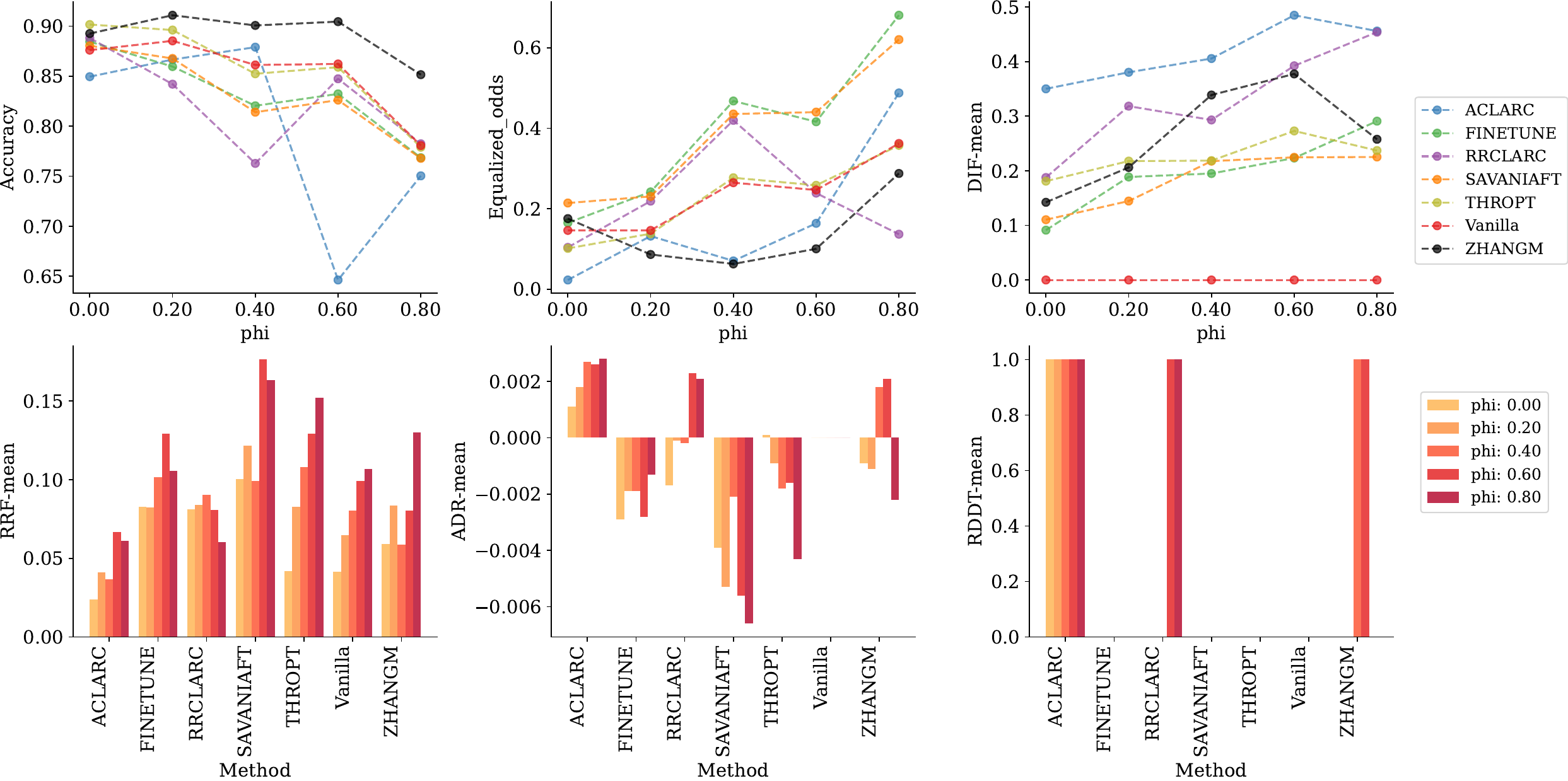}
    \caption{Metric values for the \texttt{IG} attributions and \textit{WearingHat} protected attribute}
    \label{fig:fig:IntegratedGradients-mixedplot-WearingHat}
\end{figure}

\end{document}